\pdfoutput=1

\documentclass[11pt]{article}

\usepackage[]{EMNLP2022}

\usepackage{times}
\usepackage{latexsym}

\usepackage[T1]{fontenc}

\usepackage[utf8]{inputenc}

\usepackage{microtype}

\usepackage{multirow}
\usepackage{soul}
\usepackage{color}
\usepackage{booktabs}

\definecolor{ao}{rgb}{0.0, 0.5, 0.0}
\definecolor{brickred}{rgb}{0.8, 0.25, 0.33}
\definecolor{blueish}{rgb}{0.65, 0.04, 0.37}

\usepackage{rotating}
\usepackage{makecell, multirow}

\input{webis-paper-prefix}

\begin{document}
\title{To Revise or Not to Revise: \\ Learning to Detect Improvable Claims for Argumentative Writing Support
}

\author{Gabriella Skitalinskaya\textsuperscript{1,2} \and Henning Wachsmuth\textsuperscript{1}\\
	\textsuperscript{1} Institute of Artificial Intelligence, Leibniz University Hannover \\
        \textsuperscript{2} Department of Computer Science, University of Bremen \\
	\texttt{\{g.skitalinska,h.wachsmuth\}@ai.uni-hannover.de} \\}
	
	\maketitle
\begin{abstract}
Optimizing the phrasing of argumentative text is crucial in higher education and professional development. However, assessing whether and how the different claims in a text should be revised is a hard task, especially for novice writers. In this work, we explore the main challenges to identifying argumentative claims in need of specific revisions. By learning from collaborative editing behaviors in online debates, we seek to capture implicit revision patterns in order to develop approaches aimed at guiding writers in how to further improve their arguments. We systematically compare the ability of common word embedding models to capture the differences between different versions of the same text, and we analyze their impact on various types of writing issues. To deal with the noisy nature of revision-based corpora, we propose a new sampling strategy based on revision distance. Opposed to approaches from prior work, such sampling can be done without employing additional annotations and judgments. Moreover, we provide evidence that using contextual information and domain knowledge can further improve prediction results. How useful a certain type of context is, depends on the issue the claim is suffering from, though.	
\end{abstract}

\section{Introduction}
\label{sec:introduction}

Text revision is an essential part of professional writing and is typically a recursive process until a somehow \textit{optimal} phrasing is achieved from the author's point of view. Aside from proofreading and copyediting, text revision subsumes substantive and rhetorical changes not only at the lexical, syntactic, and semantic levels, but also some that may require knowledge about the topic of discussion or about conventions of the domain or genre. An optimal phrasing is especially important in argumentative writing, where it is considered a key component in academic and professional success: An argument's style directly affects its persuasive effect on the audience \cite{el-baff-etal-2020-analyzing}.

\bsfigure{examples.ai}{Examples of revision histories of argumentative claims from the online debate platform Kialo. Colors denote claims considered optimal (light green) and claims requiring revisions (medium red). We seek to identify if and how a new claim should be revised.}

But how to know whether an argument is phrased well enough and no more revisions are needed? Most existing approaches to argument quality assessment score arguments on different aspects of a topic or compare one to another, rather than detecting issues within arguments to highlight potential improvements (see Section~\ref{sec:relatedwork} for details). Beyond those, \newcite{zhang-litman-2015-annotation} analyze the nature of revisions in argumentative writing. They annotate revisions at various levels to learn to classify changes that occur. Others compare revisions in terms of quality on essay level \cite{afrin-litman-2018-annotation} or claim level \cite{skitalinskaya-etal-2021-learning}. Still, the question of whether a given argumentative text should be revised remains unexplored. 

Figure~\ref{examples.ai} illustrates the underlying learning problem. What makes research on detecting the need~for revision challenging is the noisy and biased nature of revision-based corpora in general and respective argument corpora specifically. Not only is it uncertain whether a text will be revised in the future and how, but also the inherent subjectivity and context dependency of certain argument quality dimensions \cite{wachsmuth:2017} pose challenges. 

In this work, we investigate how to best develop approaches that identify argumentative claims in need of further revision, and that decide what type of revision is most pressing. We delineate the main challenges originating from the nature of revision-based corpora and from the notion of argument quality. 
To tackle these challenges, we exploit different types of knowledge specific to text and argument revisions: 
(a)~the number of revisions performed, in the past and the available future, 
(b)~the types of revision performed, such as typo correction vs.\ clarification, 
(c)~contextual information, such as the main thesis or parent claim in the given debate, 
(d)~topic knowledge, such as debates belonging to the same topical categories, and
(e)~the nature of revisions and their concordance with training processes of embedding representations.

In systematic experiments on a claim revision corpus, we provide evidence that some explored approaches can detect claims in need of revision well even in low-resource scenarios, if appropriate sampling strategies are used. While employing contextual information leads to further improvements in cases where linguistic cues may be too subtle, we find that it may also be harmful when detecting certain types of issues within the claim. 

We argue that technologies that identify texts in need of revision can highly benefit writing assistance systems, supporting users in formulating arguments in a better way in order to optimize their impact. The main contributions of this paper are: 
\begin{enumerate}
\setlength{\itemsep}{0pt}
\item
An overview of the main challenges in assessing whether a claim needs revision;
\item
a detailed analysis of the strengths and weaknesses of strategies to tackle said challenges, guiding future research on revisions in argumentation and other domains;
\item
a systematic comparison of approaches based on different contextualized representations for the tasks of suboptimal-claim detection and claim improvement suggestion.%
\footnote{Data, code, and models from our experiments are found at \url{https://github.com/webis-de/ACL-23}.}
\end{enumerate}

\section{Related Work }
\label{sec:relatedwork}

Foundational studies of writing specify two main revision sub-processes: evaluating (reading the produced text and detecting any problems) and editing (finding an optimal solution and executing the changes) \cite{flower1986detection}. In this work, we focus on the former in the domain of \textit{argumentative texts}. Although considerable attention has been given to the computational assessment of the quality of such texts, very few works consider the effects of revision behaviors on quality. 

Existing research largely focuses on the absolute and relative assessment of single quality dimensions, such as cogency and reasonableness \cite{marro:2022}. \newcite{wachsmuth:2017} propose a unifying taxonomy that combines 15 quality dimensions.
They clarify that some dimensions, such as acceptability and rhetorical effectiveness, depend on the social and cultural context of the writer and/or audience. A number of approaches exist that control for topic and stance \cite{habernal-gurevych-2016-argument}, model the audience \cite{alkhatib:2020}, or their beliefs \cite{el-baff-etal-2020-analyzing}, and similar. However, they all compare texts with \textit{different} content and meaning in terms of the aspects of topics being discussed. While such comparisons help characterize good arguments, they are not geared towards identifying issues within them, let alone towards guiding writers on how to improve the quality of their arguments.

The only works we are aware of that study revisions of argumentative texts are those of \citet{afrin-litman-2018-annotation}, \citet{skitalinskaya-etal-2021-learning}, and  \citet{kashefi2022argrewrite}. 
The first two suggest approaches that enable automatic assessments of whether a revision can be considered successful, that is, it improves the quality of the argumentative essay or claim. The third extends the corpus of \newcite{zhang-litman-2015-annotation} to complement 86 essays with more fine-grained annotations, enabling the distinction of content-level from surface-level revision changes at different levels of granularity.
All these approaches to characterizing the type of revision and its quality require two versions as input. In contrast, we seek to identify whether an argumentative text needs to be revised at all and, if so, what type of improvement should be undertaken. In such framing, the solutions to the tasks can also be used to support argument generation approaches, for example, by helping identify weak arguments for counter-argument generation \cite{alshomary-etal-2021-counter}, as well as automated revision approaches, for example, by providing required revision types or weak points as prompts \cite{hua-wang-2020-pair,skitalinskaya2022claim}.

Due to the lack of corpora where revisions are performed by the authors of texts themselves, researchers utilize collaborative online platforms. Such platforms encourage users to revise and~improve existing content, such as encyclopedias \cite{faruqui-etal-2018-wikiatomicedits}, how-to instructions \cite{anthonio-etal-2020-wikihowtoimprove}, Q\&A sites \cite{Li2015_collab}, and debate portals \cite{skitalinskaya-etal-2021-learning}.  
Studies have explored ways to automate content regulation, namely text simplification \cite{botha-etal-2018-learning}, detection of grammar errors \cite{lichtarge-etal-2019-corpora}, lack of citations \cite{redi2019citation}, biased language \cite{de-kock-vlachos-2022-leveraging}, and vagueness \cite{debnath-roth-2021-computational}. While \citet{bhat-etal-2020-towards} consider a task similar to ours -- detecting sentences in need of revision in the domain of instructional texts -- their findings do not fully transfer to argumentative texts, as different domains have different goals, different notions of quality, and, subsequently, different revision types performed.

Revision histories of peer-reviewed content help alleviate the shortcomings typical for self-revisions, where a writer may fail to improve a text for lack of expertise or skills \cite{fitzgerald1987research}. Yet, they also introduce new challenges stemming from sociocultural aspects, such as opinion bias \cite{garimella_2018_bias,Pryzant_2020} and exposure bias \cite{westerwick_2017_exposure}. Approaches to filtering out true positive and negative samples have been suggested to tackle such issues. These include\- community quality assessments, where high quality content is determined based on editor or user ratings and upvotes \cite{redi2019citation,chen2018data}, timestamp-based heuristics, where high-quality labels are assigned to content that has not been revised for a certain time period \cite{anthonio-etal-2020-wikihowtoimprove}, and complementary crowdsourced annotation \cite{asthana2021wikipedia}. However, all this requires domain-specific information which may not be available in general. In our experiments, we analyze the potential of sampling data solely based on revision characteristics, namely revision distance (the number of revisions between a certain claim version and its final version). 

Moreover, studies have shown that writing expertise is domain-dependent, revealing commonali\-ties within various professional and academic writing domains \cite{bazerman2016sociocultural}. Although certain quality aspects can be defined and evaluated using explicit rules, norms, and guidelines typical for a domain, not all quality aspects can be encoded in such rules. This raises the need to develop approaches capable of capturing implicit revision behaviors and incorporating additional context relevant to the decision-making process \cite{flower1986detection,boltuzic-snajder-2016-fill}.
Below, we outline the main challenges stemming from the noisy and biased nature of revision-based corpora as well as from the context dependence of certain argument quality aspects. We then establish potential data filtering and sampling strategies targeting said issues.

\section{Tasks and Challenges}
\label{sec:challenges}

Revision-based data provides many opportunities; yet, it also comes with several challenges that arise at different stages of the experiment design process. In the following, we define the tasks we deal with in this paper, summarize the main challenges, and outline our approaches to these challenges.

\subsection{Tasks}

Previous work has studied how to identify the better of two revisions. However, this does not suffice to support humans in optimizing their arguments, as it remains unclear when a claim is phrased optimally. We close this gap by studying two new tasks:

\paragraph{Suboptimal-Claim Detection} Given an argumentative claim, decide whether it is in need of further revision or can be considered to be phrased more or less optimally (binary classification).

\paragraph{Claim Improvement Suggestion} Given an argumentative claim, select all types of quality issues from a defined set that should be improved when revising the claim (multi-class classification).

\medskip \noindent
Reasons for revision can vary strongly, from the correction of grammatical errors to the clarification of ambiguity or the addition of evidence supporting the claim. In our experiments, we select quality issues sufficiently represented in the given data.

\subsection{Challenges}
\label{sec:challenge_details}

To tackle the given tasks on revision-based data, the following main challenges need to be faced:
\begin{itemize}
\setlength{\itemsep}{0pt}
\item
\emph{Data.} Compiling a dataset that is (a) represen\-tative and reliable and (b) free of topical bias.
\item
\emph{Model.} Selecting a model for the task whose complexity and architecture fit the data.
\item
\emph{Context.} Incorporating complementary contextual knowledge useful for the tasks at hand.
\end{itemize}

\noindent
We detail each challenge below and discuss how we approach it in our experiments.

\paragraph{Representativity and Reliability}  

Compiling a reliable dataset from claim revision histories that represents argumentative claim quality well is not straightforward.
While examples of suboptimal quality are rather easy to find, since each revision signals that something is wrong with the previous version, identifying examples of high (ideally, optimal) quality text remains a challenge. The main reason is that such texts remain unchanged throughout\- time in collaborative systems, yet the same holds for low-quality texts that may have been overlooked and never revised. 

Prior work largely employs external information and additional quality assessments to sample representative examples (see Section~\ref{sec:relatedwork}), limiting scalability. In this paper, we complement existing efforts by suggesting a scalable sampling strategy solely based on revision history characteristics, namely revision distance, which denotes the number of revisions that occurred until an optimal (final) state of the claim was reached. The proposed strategy as illustrated in Figure \ref{rev_distance.ai} only considers claim histories with~4 or more revisions (chosen empirically). At each revision distance $i$ from 1\,to\,4, a dataset D$i$ is compiled, where all final versions of claims are considered as positive examples not needing a revision, and claim versions~at revision distance~$i$ are considered as negative ones. 

Another problem is identifying flaws that need to be tackled in a revision. Although a claim may suffer from multiple flaws at the same time, not all of them may be eliminated in the same revision.\,In the dataset introduced in Section~\ref{sec:data}, revi\-sions may be accompanied by a label describing the type of improvement performed. Still, such~labels are skewed towards improvements addressed by the community and do not account for other flaws in the text. 

To address these issues, we explore three ways of extracting quality labels from revision histories:
\begin{itemize}
\item
We consider the revision distance between positive and negative examples when identifying claims in need of revision (Section \ref{sec:revision-depth}).
\item
We extend the given dataset with examples of claims that were never revised (Section \ref{sec:data}). 
\item
We frame the improvement suggestion task as a multi-class classification task, where only the next most probable improvement type is predicted. This better reflects the iterative nature of revision processes and accounts for the lack of explicit quality labels (Section \ref{sec:multiclass}).%
\end{itemize}

\bsfigure{rev_distance.ai}{Illustration of sampling approach. Shades of red denote claims requiring revisions at different revision distances from 1 to 4, while final versions are green and represent optimally phrased claims. }

\paragraph{Topical Bias} 

Despite the best efforts, histories of collaborative online revisions may contain noise,~be it due to accidental mistakes or biases of users or moderators. 
Different users may have conflicting opinions on what should or should not be edited, and certain debate topics may be~seen as controversial, making it even more difficult to assess the quality of claims and suggest further improvements. 

Accounting for such bias is inherently difficult and also depends on the prominence of such behaviors in the given community. We do not solve this issue here, but we explore it: 
\begin{itemize}
\setlength{\itemsep}{0pt}
\item
We determine the extent to which bias differs across topical debate categories by assessing performance differences when including claims on specific topics or not (Section \ref{sec:biases}). 
\end{itemize}

\paragraph{Model Complexity and Architecture}

Learning quality differences across several versions of the same argumentative claim likely requires a model whose architecture aligns with the idea of revisions. To determine the best model, we carry out two complementary steps:

\begin{itemize}
\setlength{\itemsep}{0pt}
\item
We train several types of models of varying complexity, including statistical and neural approaches to both obtaining claim representations and classification (Section \ref{sec:methods}).
\item
We disentangle how pre-training, fine-tuning, and the final classification affect the performance of claim assessment (Section \ref{sec:complexity}).
\end{itemize}

\paragraph{Contextuality}

As mentioned in Section \ref{sec:relatedwork}, some quality aspects require domain knowledge. However, determining what kind of information should be included when deciding whether a text needs a revision remains an open question. Some revisions\- may be typical for debate as a whole, for example\-, related to a desired structure, layout and style of citations, or choice of words for main concepts in the debate. In such cases, conditioning models\- on the debate thesis may be beneficial. Others may depend on the parent claim, which is supported\- or opposed by the claim in question, and affects whe\-ther clarifications or edits improving the relevance of the claim are needed, and potentially general domain knowledge as well \cite{gretz2020large}.
 
 Therefore, we explore contextuality as follows: 
 \begin{itemize}
\setlength{\itemsep}{0pt}
\item
We compare benefits of using contextual debate information of varying specificity when detecting suboptimal claims and recommending revision types (Sections \ref{sec:context}--\ref{sec:multiclass}). 
\end{itemize} \nopagebreak

\section{Data}
\label{sec:data}

In our experiments, we use ClaimRev \cite{skitalinskaya-etal-2021-learning}: a corpus of over 124,312 claim revision histories, collected from the online debating platform Kialo,%
\footnote{Kialo, \url{https://www.kialo.com}} 
which uses the structure of dialectical reasoning. Each debate has a main thesis, which is elaborated through pro and con claims, which allows to consider each comment as a self-contained, relevant argument. Each revision history is a set of claims in chronological order, where each claim represents a revised version of the preceding claim meant to improve a claim's quality, which holds in 93\% of all cases according to an annotation study \cite{skitalinskaya-etal-2021-learning}.

We extend the corpus by extracting 86,482 unrevised claims from the same set of debates as in ClaimRev, which have been introduced \textit{before} the reported date of data collection (June 26, 2020). Since claims that have been introduced shortly before this date are still likely to receive revisions, we additionally filter out claims that have undergone a revision within six months after the initial data collection (December 22, 2020). 
We remove all revision histories, where claim versions have been reverted to exclude potential cases of edit wars. 

Our final corpus is formed by 410,204 claims with 207,986 instances representing optimally phrased claims (positive class) and 202,218 instances requiring revisions (negative class). All claims in need of further refinement are also provided with labels indicating the specific type of improvement the claim could benefit from. In this work, we limit ourselves to the three most common types, covering 95\% of all labels revisions in the ClaimRev dataset: {clarification}, {typo/grammar} correction, and adding/correcting {links}. Specifically, \textit{clarification} means to adjust/rephrase a claim to make it more clear, \textit{typo/grammar correction} simply indicates linguistic edits, and \textit{adding/correcting links} refers to the insertion or editing of evidence in the form of links that provide supporting information or external resources to the claim. Statistics of the final dataset are shown in Table \ref{tab:data-info}. Ensuring that all versions of the same claim appear in the same split, we assign 70\% of the histories to the training set and the remaining 30\% are evenly split between the dev and test sets. 

\begin{table}[]
	\small
\renewcommand{\arraystretch}{0.95}
\setlength{\tabcolsep}{8pt}
	\centering
	\begin{tabular}{llrr}
		\toprule
		\multicolumn{1}{l}{\textbf{Subset}} &\textbf{Type}& \textbf{\# Instances} \\ \midrule
		Positive &  Final in history & 121 504 \\
		& Unrevised & 86 482  \\
		\addlinespace
		Negative & Clarification & 61 142 \\
		&Typo/Grammar & 57 219\\
		& Links & 17 467\\
		& Other/Unlabeled & 66 390  \\ \midrule
		Overall & &  410 204
		\\ \bottomrule
	\end{tabular}
	\caption{Number of instances in the extended corpus. Positive examples represent claims considered as optimally phrased. Negative examples require revisions.}
	\label{tab:data-info}
\end{table}
\section{Methods}
\label{sec:methods}

To study the two proposed tasks, we consider two experimental settings: (i) extracting claim representations by using embeddings as input to an SVM \cite{joachims-2016}, (ii) adding a classifier layer on top of pre-trained transformer models with further fine-tuning (FT). 

In our experiments, we consider the following approaches to generating claim representations: 

\begin{itemize}
\setlength{\itemsep}{0pt}
\item  
\emph{Glove} \cite{pennington-etal-2014-glove}. A static word embedding method
\item 
\emph{Flair} \cite{akbik2018coling}. A contextual character-level embedding method
\item 
\emph{BERT} \cite{devlin-etal-2019-bert}. A standard baseline pre-trained transformer
\item 
\emph{ELECTRA} \cite{Clark2020ELECTRA:}. A transformer with adversarial pre-training fitting our tasks
\item
\emph{DeBERTa} \cite{he2021deberta}. A transformer that achieved state-of-the-art performance on the SuperGLUE benchmark \cite{Wang2019Superglue}.
\end{itemize}

\section{Experiments}
\label{sec:experiments}

Based on the data from Section~\ref{sec:data} and the methods from Section~\ref{sec:methods}, we now present a series of experiments aimed at understanding the effects and possible solutions to the four challenges from Section \ref{sec:challenges}: 
(1)~the right model complexity and architecture to capture differences between claim revisions; 
(2)~representative and reliable examples of high and low quality; 
(3)~the impact of topical bias in the data; 
(4)~contextuality, where the quality of a claim may depend on its surrounding claims.

\subsection{Model Complexity and Architecture}
\label{sec:complexity}

\begin{table}[]
\small
\setlength{\tabcolsep}{3pt}
\begin{tabular}{ll@{\hspace*{-1.75em}}r@{\,\,\,}rrrr}
\toprule
\textbf{Approach}	& \textbf{Model} & \textbf{Accuracy}  & \textbf{Ma.\,F$_1$}  & \textbf{P}  & \textbf{R}  & \textbf{F$_1$}    \\ 
\midrule
\textit{Embed.}      	& Glove                                                     & 54.9 & 54.9 & 54.9          & 50.0          & 52.1 \\
\textit{+ SVM}		& Flair                                                     & 60.1 & 60.1 & 60.2          & \bf 56.9          & 58.5 \\
				& BERT                                           & 62.1 & 61.8 & 63.5          & 54.7          & 58.8 \\
				& ELECTRA                                          & 63.2 & 62.9 & 65.1          & 55.0          &  59.6 \\ 
				& DeBERTa     & 61.5 & 61.2 &  63.2 & 52.9 & 57.6                                      \\ 
\midrule
\textit{Fine-}      		&FT-BERT                                                          & 63.1 & 61.7 &  70.1          & 44.2          & 54.2 \\
\textit{tuned}      	&FT-ELECTRA                                                      &  63.8 &  62.9 &  68.8         &    49.0       & 57.2 \\ 
 & FT-DeBERTa                                                    & \bf 67.1  & \bf 66.6  &  \bf   71.3      &       55.9     &   \bf 62.6 \\ 
\midrule
\multicolumn{2}{l}{Random baseline}                                     & 50.0 & 50.0 & 50.0          & 50.0          & 50.0 \\ 
\bottomrule
\end{tabular}
\caption{Suboptimal-claim detection: Accuracy, macro F$_1$, and precision/recall/F$_1$ of the suboptimal class for all tested models, averaged over five runs. Per approach, all gains from one row to another are significant at $p < .001$ according to a two-sided student's~$t$-test.}
\label{tab:res-overall}
\end{table}

First, we explore the ability of the methods to predict whether a claim can be considered as optimal or needs to be revised. We treat all unrevised claims and final versions of claims as not needing revisions and all preceding versions as requiring revisions. 

Table \ref{tab:res-overall} presents the results of integrating several embeddings with a linear SVM classifier and fine-tuned transformer-based language models. Although we see gradual substantial improvements as we increase the complexity of the models used to generate the word embeddings, the best results (accuracy 67.1, macro F$_1$ 66.6) indicate the difficult nature of the task itself. Low results of \emph{Glove} (both 54.9) indicate that general word co-occurrence statistics are insufficient for the task at hand. And while switching to contextualized word embeddings, such as \emph{Flair}, leads to significant improvements, pre-trained transformers perform best. 

The difference between the transformer-based models suggests that the pre-training task and attention mechanism of models impact the results notably. Unlike BERT, \emph{ELECTRA} is pretrained on the replaced-token detection task, mimicking certain revision behaviors of human editors (e.g., replacing some input tokens with alternatives). Using ELECTRA boosts accuracy from 62.1 to 63.2 for non-finetuned models and from 63.1 to 63.8 for fine-tuned ones. 
\emph{FT-DeBERTa} further improves the accuracy to 67.2, suggesting that also separately encoding content and position information, along with relative positional encodings, make the model more accurate on the given tasks. We point out that, apart from considering alternative pre-training strategies, other sentence embeddings and/or pooling strategies may further improve results.

\paragraph{Error Analysis} 

Inspecting false predictions revealed that detecting claims in need of revisions concerning corroboration (i.e., \textit{links}) is the most challenging (52\% of such cases have been misclassified). This may be due to the fact that corroboration examples are underrepresented in the data (only 13\% of the negative labeled samples). Accordingly, increasing the number of training samples could lead to improvement. In the appendix, we provide examples of false negative and false positive predictions. They demonstrate different cases where claims are missing necessary punctuation, clarification, and links to supporting evidence.

\subsection{Representativeness and Reliability}
\label{sec:revision-depth}

Next, we explore the relationship between revision distance and data reliability by using the sampling strategy proposed in Section \ref{sec:challenges}. We limit our experiments to revision histories with more than four revisions and compile four datasets, each representing a certain revision distance. We use the same data split as for the full corpus, resulting in 11,462 examples for training, 2554 for validation, and 2700 for testing for each of the sampled datasets. 

\begin{table}[]
\small
\centering
\renewcommand{\arraystretch}{1}
\setlength{\tabcolsep}{5pt}
\begin{tabular}{lrrrrr}
\toprule
\multicolumn{1}{l}{{\textbf{Training Subset}}} & \textbf{D1} & \textbf{D2} & \textbf{D3} & \textbf{D4} & \textbf{Average}\\ 
\midrule
{D1} & 53.8 & 56.3 & 58.1 &     60.5      & 57.2  \\
{D2} & 55.1 & 58.4 &  60.1  & 63.6 & 59.4 \\
{D3} & 55.2 & 58.4 &  61.1&   64.2   &  59.7 \\
{D4} & \textbf{55.5}& \textbf{59.0} & \textbf{61.8} & \textbf{65.6} &  \textbf{60.5} \\ 
\midrule
{Full training set} & 56.8 & 61.2 & 64.3 & 65.8 & 62.0 \\ 
\bottomrule
\end{tabular}
\caption{Accuracy of FT-ELECTRA averaged over five runs, depending on sample training subset and test subset D$i$; $i$ denotes the revision distance from 1 to~4.}
\label{tab:res-distance}
\end{table}

Table \ref{tab:res-distance} shows the accuracy scores obtained by \emph{FT-ELECTRA}, when trained and tested on each sampled subset D$i$. Not only does the accuracy increase when training on a subset with a higher revision distance (results per column), but also the same model achieves higher accuracy when classifying more distant examples (results per row). On one hand, this indicates that, the closer the claim is to its optimal version, the more difficult it is to identify flaws. On the other hand, when considering claims of higher revision distance, the model seems capable of distinguishing optimal claims from other improved but suboptimal versions. 

Comparing the results of training on D4 with those for the full training set, we see that the D4 classifier is almost competitive, despite the much smaller amount of data. For example, the accuracy values on the D4 test set are 65.6 and 65.8, respectively. We conclude that the task at hand can be tackled even in lower-resource scenarios, if a representative sample of high quality can be obtained. This may be particularly important when considering languages other than English, where argument corpora of large scale may not be available.

\subsection{Topical Bias}
\label{sec:biases}

To measure the effects of topical bias, in a first experiment we compare the accuracy per topic category of \emph{FT-DeBERTa} and \emph{FT-ELECTRA} to detect whether identifying suboptimal claims is more difficult for certain topics. In Table \ref{tab:res-topic}, we report the accuracy for the 20 topic categories from \newcite{skitalinskaya-etal-2021-learning}. We find that the task is somewhat more challenging for debates related to topics, such as \emph{justice}, \emph{science}, and \emph{democracy} (best accuracy 63.6--65.2) than for \emph{europe} (69.1) or \emph{education} (68.9). We analyzed the relationship between the size of the categories and the models' accuracy, but found no general correlation between the variables indicating that the performance difference does not stem from how represented each topic is in terms of sample size.%
\footnote{A scatter plot of size vs.\ accuracy is found in the appendix.}

In a second experiment, we evaluate how well the models generalize to unseen topics. To do so, we use a leave-one-category-out paradigm, ensuring that all claims from the same category are confined to a single split. We observe performance drops for both \emph{FT-DeBERTa} and \emph{FT-ELECTRA} in Table~\ref{tab:res-topic}. The differences in the scores indicate that the models seem to learn topic-independent features that are applicable across the diverse set of categories, however, depending on the approach certain topics may pose more challenges than others, such as \emph{religion} and \emph{europe} for \emph{FT-DeBERTa}.

\begin{table}[]
\small
\centering
\renewcommand{\arraystretch}{0.9}
\setlength{\tabcolsep}{8pt}
\begin{tabular}{l@{}rcrc}
\toprule
& \multicolumn{2}{r}{\bf FT-ELECTRA}	& \multicolumn{2}{c}{\bf FT-DeBERTa}  \\ \cmidrule(l@{2pt}r@{2pt}){2-3}			\cmidrule(l@{2pt}r@{2pt}){4-5}	
\textbf{{Category}} & \quad \bf Full & \bf Across & \bf Full & \bf Across \\
\midrule
     Education &     67.0 & 66.2 & \bf 68.9 & 68.6 \\
    Technology &    65.7 &64.3 & 66.9 & {\bf 67.0} \\
    Philosophy &     65.0 &  65.2 & \bf 67.3 & 67.1 \\
        Europe &     65.3 & 64.6 & \bf 69.1 & 66.5 \\
     Economics &     64.8 & 65.1 & \bf 68.0 &  66.2\\
     \addlinespace
    Government &     65.2 & 64.7  & \bf 67.7 & 66.8  \\
           Law &     64.5 & 62.0  &\bf 67.7 & 65.9 \\
        Ethics &     64.7 & 64.4  &\bf 67.4 & 66.1 \\
      Children &     64.2 & 62.0   & \bf 67.2 & 66.0 \\
       Society &     64.5 & 63.6  & \bf 67.1 & 66.2 \\
       \addlinespace
        Health &     65.0 & 64.7   & \bf 68.7 & 66.5 \\
      Religion &     64.2 & 63.9  & \bf 67.5 & 63.4 \\
        Gender &     63.4 & 62.9  & \bf 66.8 & 65.0 \\
 ClimateChange &     63.2 & 62.8  & \bf 66.0 & 63.8 \\
      Politics &     62.6 & 62.2 &\bf 66.5 & 64.7 \\
      \addlinespace
           USA &     62.0 & 62.2  & \bf 65.4 & 64.0 \\
       Science &     61.9 & 61.0  & \bf 65.2 & 62.8 \\
       Justice &     60.2 & 58.6  & \bf 63.6 & 61.2 \\
      Equality &     62.9 & 61.2  & \bf 67.5 & 65.5 \\
     Democracy &     61.3 & 60.3  & \bf 65.2 & 63.4 \\
\bottomrule
\end{tabular}
\caption{Topical bias: Accuracy of FT-ELECTRA an FT-DeBERTa across 20 topic categories, when trained on the full dataset (\emph{full}) and in a cross-category setting using a leave-one-out strategy (\emph{across}).}
\label{tab:res-topic}
\end{table}

Overall, the experiments indicate that the considered approaches generalize fairly well to unseen topics, however, further evaluations are necessary to assess whether the identified topical bias is due to the inherent difficulty of certain debate topics, or the lack of expertise of participants on the subject resulting in low quality revisions, requiring the collection of additional data annotations.

\subsection{Contextuality}
\label{sec:context}

In our fourth experiment, we explore the benefits of incorporating contextual information. We restrict our view to the consideration of the main thesis and the parent claim, each representing context of different levels of broadness. We do so by concatenating the context and claim vector representations in SVM-based models, and by prepending the context separated by a delimiter token when fine-tuning transformer-based methods.

Table \ref{tab:res-context} reveals that, overall, adding context leads to improvements regardless~of the method used. Across all approaches, including the narrow context of the parent claim seems more important for identifying suboptimal claims, with the best result obtained by \emph{FT-DeBERTa} (accuracy of 68.6). 

The results also suggest that classification approaches employing non-finetuned transformer embeddings and contextual information can achieve results comparable to fine-tuned models, specifically models with a high similarity of the pretraining and target tasks \cite{peters-etal-2019-tune}. However, some quality aspects may require more general world knowledge and reasoning capabilities, which could be incorporated by using external knowledge bases. We leave this for future work.

\begin{table}[t]
\centering
\small
\setlength{\tabcolsep}{4pt}
\begin{tabular}{l@{}rrrrr}
	\toprule
\textbf{Model} & \textbf{Accuracy}  & \textbf{Ma.\,F$_1$}  & \textbf{P}  & \textbf{R}  & \textbf{F$_1$}    \\ 
\midrule
Glove+SVM&
  54.9 &
  54.9 &
  54.9 &
  50.0 &
  52.1
  \\
 \qquad + thesis &
  55.9 &
  55.8 &
  55.6 &
  53.1 &
  54.3 
  \\
 \qquad + parent &
  56.9 &
  56.9 &
  56.3 &
  57.3 &
  56.8 
  \\
\addlinespace
Flair+SVM&
  60.1 &
  60.1 &
  60.2 &
  56.9 &
  58.5 
  \\
\qquad + thesis &
  62.4 &
  62.4 &
  62.0 &
  61.4 &
  61.7
  \\
 \qquad + parent &
  62.8 &
  62.8 &
  61.9 &
  \bf 64.4 &
  63.1
  \\
\addlinespace
BERT+SVM&
  62.1 &
  61.8 &
  63.5 &
  54.7 &
  58.8 
  \\
 \qquad + thesis &
  63.5 &
  63.4 &
  64.2 &
  59.0 &
  61.5  
  \\
 \qquad + parent &
  63.8 &
  63.8 &
  64.0 &
  61.0 &
  62.5 
  \\
\addlinespace
ELECTRA+SVM&
  63.2 &
  62.9 &
  65.1 &
  55.0 &
  59.6 
  \\
 \qquad + thesis  &
  65.0 &
  64.9 &
  66.1 &
  60.0 &
  62.9 
  \\ 
 \qquad + parent &
  65.2 &
  65.1 &
  65.4 &
  62.6 &
  64.0
  \\ 
  \addlinespace
DeBERTa+SVM&
    61.5 & 61.2 &  63.2 & 52.9 & 57.6   
  \\
 \qquad + thesis  & 62.5 & 62.2 & 63.9 & 55.1 & 59.2
  \\ 
 \qquad + parent & 63.3 & 63.2 & 64.0 & 59.0 & 61.4
  \\ 
\addlinespace
FT-BERT &
  63.1 &
  61.7 &
  70.1 &
  44.2 &
  54.2 
  \\
 \qquad + thesis &
  64.1 &
  63.0 &
  70.1 &
  47.6 &
  56.7 
  \\
 \qquad + parent &
 65.7 &
  65.4 &
  67.5 &
  58.8 &
  62.8 
  \\
\addlinespace
FT-ELECTRA &
 63.8 &
  62.9 &
  68.8 &
  49.0 &
  57.2 
   \\
 \qquad + thesis &
  64.4 &
  63.5 &
  69.2&
  50.4&
  58.2
   \\ 
 \qquad + parent &
  64.8 &
  64.6 &
  66.0 &
  59.3 &
  62.4 

 \\
\addlinespace
 FT-DeBERTa & 67.1  &  66.6  &     71.3      &       55.9     &    62.6

   \\
 \qquad + thesis &  67.3
  &  67.0
   &  70.1
  & 59.5
  & 64.2

   \\ 
 \qquad + parent &  \bf 68.6
 & \bf 68.4
 &  \bf 71.4
 &  60.8
 &  \bf 65.7

  \\ \midrule
Random baseline &
  0.50 &
  0.50 &
  0.50 &
  0.50 &
  0.50 
   \\ \bottomrule
\end{tabular}
\caption{Contextuality: Results of all evaluated models when including the \emph{thesis} or \emph{parent} as contextual information, averaged over five runs. For each approach, all gains from one row to another are significant at $p < .001$ according to a two-sided student's $t$-test.}
\label{tab:res-context}
\end{table}

\subsection{Claim Improvement Suggestion}
\label{sec:multiclass}

\begin{table}[]
\small
\setlength{\tabcolsep}{4pt}
\begin{tabular}{l@{\hspace*{-1.5em}}rrrrr}
\toprule
 &
 &
&
  \multicolumn{3}{c}{\textbf{F$_1$-Score}}
  \\ \cmidrule{4-6}
\textbf{Setup}   &
  \textbf{Accuracy}   &
  \textbf{Ma.\,F$_1$}    &
  \textbf{Clarif.} &
  \textbf{Typo} &
  \textbf{Links} \\ 
 \midrule
FT-ELECTRA & 56.0 & 49.0 & 62.4 & 52.4 &  34.5\\
 \qquad + parent   & 56.2
  & 50.3
  & 62.0
  & 53.6
  & 35.3
  \\
 \qquad + thesis  & 57.5
  & 52.0
  & 63.4
  & 54.4
& 38.3\\
\addlinespace
 FT-DeBERTa & 59.9 & 55.4 & 63.7 &  60.2 &   42.5 \\
 \qquad + parent   &  60.3
   &   56.0
   &   63.6
   &  61.2
   &  43.0
   \\
 \qquad + thesis  &  \bf 62.0
   &  \bf 57.3
   &  \bf 65.2
   & \bf  63.1
&\bf 43.4
   \\ \midrule
{{Random baseline}}  &  33.4 & 31.4 & 38.5 & 33.4 & 45.3 
\\ \bottomrule
\end{tabular}
\caption{Claim improvement suggestion: Accuracy, macro F$_1$-score, and the F$_1$-score per revision type for ELECTRA+SVM and FT-DeBERTa with and without considering context, averaged over five runs.}
\label{tab:res-type}
\end{table}

While previous experiments have shown the difficulty of distinguishing between claims in need of improvements and acceptable ones, the aim of this experiment is to provide benchmark models for predicting the type of improvement that a certain claim could benefit from. Here, we limit ourselves to the three most common types of revision: \emph{clarification}, \emph{typo and grammar correction}\- (includes style and formatting edits), and \emph{adding/correcting links} to evidence in the form of external sources. Additional experiments covering an end-to-end setup by extending the classes to include claims that do not need revisions can be found in the appendix.   
We compare two best performing models from previous experiments, FT-ELECTRA and FT-DeBERTa, on a subset of~135,828~claims, where editors reported any of the three types.

Table \ref{tab:res-type} emphasizes the general benefit of utilizing contextual information for claim improvement suggestion. Though, depending on the specific revision type, the addition of contextual information can both raise and decrease performance. For example, despite the slightly improved overall accuracy of considering the parent claim as context, the F$_1$-score for \textit{clarification} edits drops for all considered approaches (from 63.7 to 63.6 for FT-DeBERTa and from 62.4 to 62.0 for FT-ELECTRA). On the other hand, in the case of \textit{links}, both types of contextual information lead to increased F$_1$-scores. Generally, we notice that opposed to the task of suboptimal-claim detection, providing the main thesis of the debate leads to higher score improvements overall.
When comparing the approaches directly, we observe that \emph{FT-DeBERTa} consistently outperforms \emph{ELECTRA+SVM} in accuracy, achieving 62.0 when considering the main thesis.

Overall, our experiments indicate that to identify whether certain approaches to generating text representations are more suitable than others, one needs to consider the relationships between improvement type and context as well. In future work, we plan to focus on the problem of further defining and disentangling revision types to enable a deeper analysis of their relationships with contextual information.

\paragraph{Error Analysis} 

Inspecting false predictions of the best performing model (FT-DeBERTa) revealed that the typo/grammar correction class seems to be confused frequently with both the clarification class and the links class (see the appendix for a confusion matrix). Our manual analysis suggests that editors frequently tackle more than one quality aspect of a claim in the same revision, for example, specifying a claim and fixing grammatical errors, or, removing typos from a link snippet. In the collected dataset, however, the revision type label in such cases would reflect only one class, such as \emph{clarification} or \emph{adding/correcting links}, respectively. These not fully accurate labels reduce the models ability to properly distinguish said classes. We provide examples of misclassifications obtained by FT-DeBERTa in the appendix, illustrating cases where both the true label and the predicted label represent plausible revision type suggestions. 
\section{Conclusion}

Most approaches to argument quality assessment rate or compare argumentative texts that cover different aspects of a topic. While a few works studied which of two revisions of the same argumentative text is better, this does not suffice to decide whether a text actually needs revisions. 

In this paper, we have presented two tasks to learn \emph{when} and \emph{how} to improve a given argumentative claim. We have delineated the main challenges of revision-based data, covering issues related to the representativeness and reliability of data, topical bias in revision behaviors, appropriate model complexities and architectures, and the need for context when judging claims. In experiments, we have compared several methods based on their ability to capture quality differences between different versions of the same text. Despite a number of limitations (discussed below), our results indicate that, in general, revision-based data can be employed effectively for the given tasks, contributing towards solutions for each of the considered challenges. Specifically, our suggested sampling strategy revealed that training on claim versions with a higher revision distance between them improves the performance when detecting claims in need of improvement. Moreover, we found that the impact of the available types of contextual information is not only task-dependent but also depends on the quality issue that a claim suffers from.

We argue that the developed approaches can help assist automated argument analysis and guide writers in improving their argumentative texts. With our work, we seek to encourage further research on improving writing support not only in debate communities but in educational settings as well.
\section*{Acknowledgments}
We thank Andreas Breiter for his valuable feedback on early drafts, and the anonymous reviewers for their helpful comments. This work was partially funded by the Deutsche Forschungsgemeinschaft (DFG, German Research Foundation) under project number 374666841, SFB 1342. 
\section*{Limitations}

A limitation of our work is that we cannot directly apply our methods to the few existing revision-based corpora from other domains \cite{yang-etal-2017-identifying-semantic,afrin-litman-2018-annotation,anthonio-etal-2020-wikihowtoimprove} for multiple reasons: On the one hand, those corpora do not contain histories with more than one revision but only before-after sentence pairs). Some also consist of less than 1000 sentence pairs, rendering the quantitative experiments considered in this paper pointless. On the other hand, additional metadata useful for our analysis (e.g., revision types and contextual information) is either not available at all or only for a limited number of instances that is insufficient for training models.

Furthermore, the methods we evaluated utilize distantly supervised labels based on the assumption that each revision improves the quality of the claim and additional annotations provided by human editors. These annotations suffer from being coarse-grained, consisting of mainly three classes. However, each of the improvement types can be represented by several more fine-grained revision intentions. 
A point that we did not consider as part of this work is whether certain revisions can affect or inform future revisions within the same debate, for example, rephrasing of arguments to avoid repetition or ensuring that all claims use the same wording for the main concepts. Often, such relationships are implicit and cannot be derived without additional information provided by the user performing the revision. We believe that collecting datasets and developing approaches, which enable distinguishing more fine-grained types of edits and implicit relationships, could not only enable deeper analysis and training more fine-grained improvement suggestion models, but also allow for better explanations to end users.

However, it should be noted that some of the considered methods rely on deep learning and have certain limitations when it comes to underrepresented classes, where the number of available training instances is very low. This is especially important when considering the task of claim improvement suggestion. We also point out in this regard that we only use the base versions of the BERT, ELECTRA, and DeBERTa models due to resource constraints. The results may vary, if larger models are used.  

While common types of improvements likely differ across other domains and communities, we stress that our approaches are entirely data-driven, and are not tied to any specific quality definition. Therefore, we expect our data processing and filtering methods as well as the considered approaches to be applicable to other domains, where historical collaborative editing data similar to ours is available. When it comes to practice, several issues require further investigation, such as how to integrate recommendations in collaborative editing and educational environments, whether the recommended improvements will be accepted by users, and how they may impact the users' behavior. We leave these questions for future work.

\section*{Ethical Considerations}

Online collaborative platforms face challenging ethical problems in maintaining content quality. On the one hand, they need to preserve a certain level of free speech to stimulate high quality discussions, while implementing regulations to identify editing behaviors defined as inappropriate. On the other hand, distinguishing such legitimate forms of regulation from illegitimate censorship, where particular opinions and individuals are suppressed, is a challenge of its own.

Our work is intended to support humans in different scenarios, including the creation or moderation of content on online debate platforms or in educational settings. In particular, the presented approaches are meant to help users by identifying argumentative claims in need of further improvements and suggesting potential types of improvements, so they can deliver their messages effectively and honing their writing skills. However, the presented technology might also be subject to intentional misuse, such as the above-mentioned illegitimate censorship. While it is hard to prevent such misuse, we think that the described scenarios are fairly unlikely, as such changes tend to be noticed by the online community quickly. Moreover, the source of the used data (online debate platform Kialo) employs thorough content policies and user guidelines aimed at dealing with toxic behaviors, censorship, and discrimination. However, we suggest that follow-up studies stay alert for such behaviors and carefully choose training data.

\bibliography{anthology,custom}
\bibliographystyle{acl_natbib}

\clearpage
\appendix

\section{Implementation and Training Details}
\subsection{Generating Embeddings}

All claim embeddings were generated using the flair library,%
\footnote{flair, \url{https://github.com/flairNLP/flair}}
via DocumentPoolEmbeddings for non-transformer-based models, such as Glove and Flair, or TransformerDocumentEmbeddings for BERT and ELECTRA embeddings.

\paragraph{Glove + SVM} We derived claim representations by averaging the obtained word representations and feed them as input to a linear SVM \cite{joachims-2016}. We initialized the 100-dimensional word embeddings pretrained on Wikipedia data ("glove-wiki-gigaword-100"). 

\paragraph{Flair + SVM} We used the 2,048-dimension ``news-forward'' embeddings, produced by a forward bi-LSTM, trained on the One Billion Word Benchmark \cite{chelba2013one} and feed the obtained embeddings to a linear SVM classifier. 

\paragraph{BERT}  We use the case-sensitive pre-trained version (bert-base-cased).

\subsection{Training SVM models}

For faster convergence when dealing with a large number of samples, we use a SVM with a linear kernel, specifically, LinearSVC, as implemented in the sklearn library.%
\footnote{sklearn SVM, \url{https://scikit-learn.org/stable/modules/generated/sklearn.svm.LinearSVC.html\#sklearn.svm.LinearSVC}} 
We set maximum iterations to 1000 and choose the regularization parameter out of \{0.001, 0.01, 0.1, 1, 10\}.

\subsection{Fine-tuning Transformer-based models}
We used the \textit{bert-base-cased} pre-trained BERT version (110M parameters), the \textit{electra-base-discriminator} pre-trained ELECTRA version (110M parameters), and the \textit{deberta-base} pretrained DeBERTA version (140M parameters) as implemented in the huggingface library.%
\footnote{Huggingface transformers, \url{https://huggingface.co/transformers/pretrained\_models.html}}
We set the maximum sequence length to 128 and 256 tokens, depending whether contextual information was used or not. We trained for a maximum of five epochs using the Adam optimizer with a warmup of 10000 steps and linear learning rate scheduler. We chose the learning rate out of \{5e-7, 1e-6, 5e-6, 1e-5, 5e-4\} and found that 1e-5 works best for BERT and DeBERTa, and 1e-6 -- for ELECTRA. In all experiments, the batch size was set to 8.  The training time on one RTX 2080Ti GPU was 80--160 minutes, depending on the chosen setup (with or without context information).

\subsection{Data and Models}
All dataset extensions and trained models are available under the CC-BY-NC license.

\section{Prediction Outputs}

\subsection{Suboptimal Claim Detection}

Table \ref{tab:error_examples} provides examples of false negative and false positive predictions obtained by FT-DeBERTa (without considering context) illustrating common patterns found in the results. 

\subsection{Claim Improvement Suggestion}

Table \ref{tab:confusion_multi} presents the confusion matrix of predictions made by FT-DeBERTa (without considering context) illustrating misclassification patterns found in the results.

Table \ref{tab:error_examples_multi} provides examples of misclassifications obtained by the best performing model (FT-DeBERTa), illustrating cases where both the true class label and the predicted class label represent plausible revision type suggestions.

\subsection{End-to-end Setup}

Table \ref{tab:res-type-app} provides extended performance results obtained by approaches using ELECTRA and DeBERTA in an end-to-end setup, where both optimal claim detection and improvement suggestion tasks are combined into one multiclass classification task with four classes: \textit{optimal} (claim does not need revisions), needs \textit{clarification}, needs \textit{typo} and/or grammar correction, needs editing of \textit{links}. 

The results suggest that in such setup it is highly difficult to detect claims requiring clarification edits (F1-scores of 15.3 (\textit{FT-DeBERTa} with parent) and 1.5 (\textit{FT-ELECTRA} with parent). Such low scores can be partially explained by (a) the high diversity of changes included in the class compared to \textit{typo} and \textit{links} classes, (b) the high imbalance of the data (percentage of samples per class: clarification~(18\%), typo~(17\%), links~(5\%), and optimal~(60\%)).

Table \ref{tab:res-type} emphasizes the general benefit of utilizing contextual information, however, similar to the results obtained in the task of claim improvement suggestion, depending on the specific revision type, the addition of contextual information can both raise and decrease performance. Particularly, we observe decreased performance in \textit{FT-DeBERTa} when detecting \textit{clarifications} and \textit{link} corrections while considering the parent claim as context. On the other hand, in the case of \textit{typo/grammar} and \textit{optimal} claims, both types of contextual information lead to increased F$_1$-scores. Generally, we notice that similar to the task of claim improvement suggestion, providing the main thesis of the debate leads to higher score improvements overall.

As indicated previously, further defining and disentangling revision types along with their relationships to contextual information could further benefit not only our understanding of revision processes in argumentative texts and their relationship to quality, but also help overcome modeling limitations identified in this paper. 

\begin{table}[t]
\small
	\centering
\renewcommand{\arraystretch}{1}
\setlength{\tabcolsep}{3pt}
\settowidth\rotheadsize{\theadfont Trueth}
	\begin{tabular}{@{\hspace*{-0.5em}}rrrrp{2em}}
	&  \multicolumn{3}{c}{\textit{Predicted}}  \\
		\cmidrule[1pt]{2-4}
		 &\textbf{Clarification}& \textbf{Typo} & \textbf{Links} \\ \cmidrule{2-4}
		\bf Clarification &  5884 (.64) & 2593 (.28) & 709 (.08) & \multirow{3}{*}[1ex]{\rothead {\textit{True}}}\\
		\bf Typo & 2788 (.33) & 5214 (.61) & 483 (.06) \\
		\bf Links & 1020 (.39) & 544 (.21)  & 1067 (.41)
		\\ \cmidrule[1pt]{2-4}
	\end{tabular}
	\caption{Claim improvement suggestion: Confusion matrix obtained by FT-DeBERTa without using context. }
	\label{tab:confusion_multi}
\end{table}
\begin{table}[]
\small
\renewcommand{\arraystretch}{0.85}
\setlength{\tabcolsep}{0.7pt}
\begin{tabular}{l@{\hspace*{-1.5em}}rrrrrrr}
\toprule
 &
 &
&
  \multicolumn{4}{c}{\textbf{F$_1$-Score}}
  \\ \cmidrule{4-7}
\textbf{Setup}   &
  \textbf{Accuracy}   &
  \textbf{Ma.\,F$_1$}    &
  \textbf{Clarif.} &
  \textbf{Typo} &
  \textbf{Links} & 
  \textbf{Optimal} &\\ 
 \midrule

FT-ELECTRA & 62.7 & 32.4 & 0.0 & 33.6 &  19.1 & 76.8\\
 \qquad + parent   & 62.9
  & 33.0
  & 0.0
  & 33.5
  & 21.2 &77.1
  \\
 \qquad + thesis  & 63.3
  & 34.1
  & 1.5
  & 36.8
& 20.7 & 77.3\\
\addlinespace
 FT-DeBERTa & 64.2 & 39.8 & 9.4 &  43.0 &   28.9 & 78.0 \\
 \qquad + parent   &  64.8
   &   40.3
   &   9.1
   &  45.4
   &  28.1 & 78.5
   \\
 \qquad + thesis  &  \bf 65.5
   &  \bf 42.7
   &  \bf 15.3
   &  47.0
&\bf 29.6 & 78.8
   \\ \midrule
{{Random baseline}}  &  25.0 & 21.1 & 20.8 & 19.8 & 8.4 & 35.5
\\ \bottomrule
\end{tabular}
\caption{Combining Improvement Suggestion and Optimal Claim Detection: Accuracy, macro F$_1$-score, and the F$_1$-score per revision type for ELECTRA+SVM and FT-DeBERTa with and without considering context, averaged over five runs.}
\label{tab:res-type-app}
\end{table}
\begin{table*}[]
\small
\centering
\renewcommand{\arraystretch}{1}
\setlength{\tabcolsep}{5pt}
\begin{tabular}{p{0.42\textwidth}p{0.42\textwidth}}
\toprule
\textbf{False Positives}  & \textbf{False Negatives} \\ \midrule
The HPV virus is harmful. (Clarif) & can be dangerous for bikers\\
\addlinespace
Vertically farming is healthier for people. (Clarif) &Women are healthier than men   \\
\addlinespace
There would be disputed over the leaders  (Typo/Grammar) & I can't support this. The math is way off. We have 15X the population and 55X the homicide rate. \\
\addlinespace
The world is becoming too populated anyway.  (Style)  &
People are likely to forget distressing memories.  \\
\addlinespace
The Czech Republic is funding travel TV shows in Korea.  (Links) &The police of every country have abused their authority systemically at some point in history \\
\addlinespace
A number of recreational drugs may have health benefits. (Links) & Podcasts cannot include music due to copyright issues, so they cannot replace radio entirely 
 \\ 
 
\bottomrule
\end{tabular}
\caption{Examples of False Positive and False Negative predictions obtained by FT-DeBERTa (without considering context). The true class for False Positives is reflected in the brackets at the end of each claim. }
\label{tab:error_examples}
\end{table*}

\section{Figures}

\subsection{Topical Categories}

\begin{figure*}[]
	\centering
	\includegraphics[width=\linewidth]{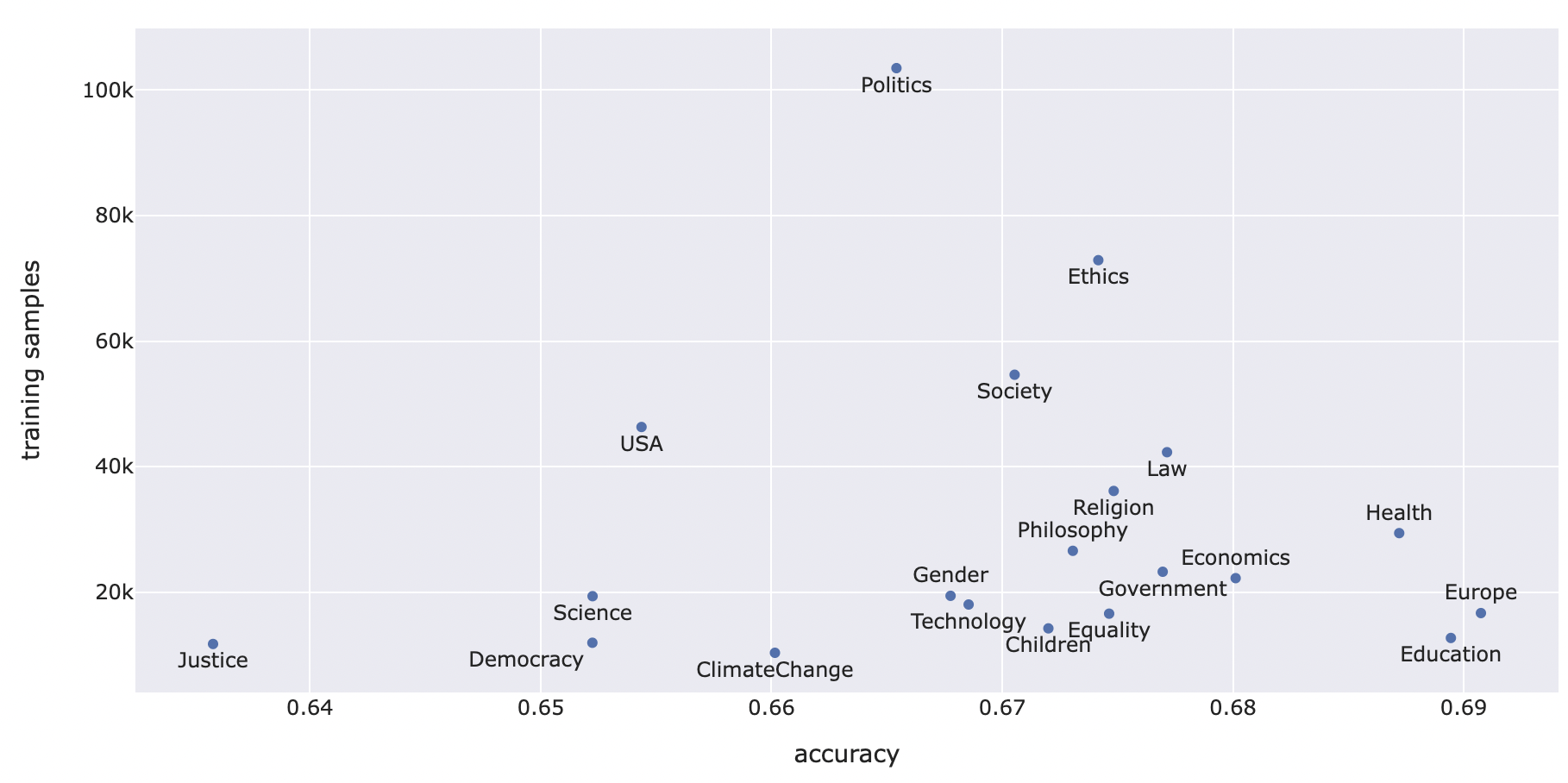}
	\caption{ Scatter plot of training sample size vs. accuracy
for 20 topical categories of the extended ClaimRev corpus achieved by FT-DeBERTa in the cross-category setting.} 
	\label{fig:scatter}
\end{figure*}

Figure \ref{fig:scatter} depicts the relationship between how represented the topical category is in the corpus and the achieved prediction accuracy by FT-ELECTRA in the cross-category setting using a leave-one-out-strategy.

\begin{table*}[]
\small
\centering
\renewcommand{\arraystretch}{1}
\setlength{\tabcolsep}{5pt}
\begin{tabular}{p{0.60\textwidth}rr}
\toprule
\textbf{Claim}  & \textbf{True Label} & \textbf{Predicted Label} \\ \midrule
\addlinespace

Freedom of speech is exceptionally good in the US, despite a recent decline in its acceptance & clarif & links \\ 
\addlinespace
Muslim women must remove their burkas for their driver's license.& clarif & links \\
\addlinespace
Voluntary help is beneficial to Germany & clarif & gram \\
\addlinespace
indecent exposure violated the right of free expression, and is therefore an illegal law.& clarif & gram \\
\addlinespace
Public restrooms should be gender neutral. & clarif & gram\\ 
\midrule
\addlinespace

Not all platforms aid terrorists' cause. Those who do not will not be censored or shut down. & typo & clarif \\
\addlinespace
The use of nuclear weapons was required in order to end the Pacific War between the US and Japan.& typo & clarif \\
\addlinespace
Nuclear weapons have spread to politically unstable states, for example Pakistan which experienced stagflation during the 1990s, a military coup in 1999 as well as a unsuccessful coup attempt in 1995. & typo &  links \\
\addlinespace
Many of the animals are now extinct, such as mammoths, mastodons, aurochs, cave bears ect. & typo & links \\
\midrule
\addlinespace

For example, the one who will have more than one wife, should equally treat all his wives.[Link](http://islamqa.info/en/14022) & links &  clarif\\
\addlinespace
Before the nuclear bombs were dropped 70\% of suitable targets had already been completely destroyed by conventional bombing. & links &  typo \\
\addlinespace
For the Spanish bullfighting is a way to reconnect to old, traditional and great Spain and therefore a major source of identity. & links & gram \\
\addlinespace
DDOS attacks are the online equivalent of a sit-in.& links &  clarif\\
 
\bottomrule
\end{tabular}
\caption{Examples of misclassifications obtained by TF-DeBERTa (without considering context).}
\label{tab:error_examples_multi}
\end{table*}

\end{document}